# Multi-vision Attention Networks for On-line Red Jujube Grading[†]

SUN Xiaoye[1,2], MA Liyan[3*] and LI Gongyan[1]

1. Institute of Microelectronics of Chinese Academy of Sciences, Beijing 100029, China
2. University of Chinese Academy of Sciences, Beijing 100049, China
   Email: sunxiaoye16@mails.ucas.ac.cn
3. School of Computer Engineering and Science, Shanghai University, Shanghai 200444, China
   Email: liyanma@shu.edu.cn

**Abstract** — To solve the red jujube classification problem, this paper designs a convolutional neural network model with fast speed, small structure and high classification accuracy. The structure of the model is inspired by the multi-visual mechanism of the organism and designed based on the DenseNet architecture. To further improve the ability of our model, we add the attention mechanism of SE-Net. To construct the data set, we capture 23,735 red jujube images via a jujube grading system. According to the appearance of the jujube surface and the feature of the sorting system, the data set are divided into four classes: invalid, rotten, wizened and normal. The numerical experimental results show that our network model can achieve a classification accuracy of 91.89%, which is comparable to DenseNet121，InceptionV3 , V4 networks, and Inception-ResNet v2. However, our model achieves the real-time performance, about 2.21ms per image.

## I. Introduction

The red jujube used as food and herbs has a history of more than 3,000 years in China [1], where is currently the world's largest producer and exporter of jujube. The homogeneity and appearance of fruits and vegetables have significant impact on consumer decision [2]. Fruits and vegetables with better appearance can be sold at high prices. Thus, the classification of red jujubes can improve the economic benefits of producers, especially in China, the largest red jujube country.

At present, most of the fruit classification algorithms are based on traditional image processing algorithms, which need the hand-crafted features for different situations. To design those features, one would take a lot of time and effort [3].

In recent years, with the progress of deep learning technology, image classification obtained great improvements. For fruit grading and classification, deep learning method is more powerful than traditional image processing algorithms [3]. It is good at feature extraction and representation, especially for automatically extracting features from raw data [4]. And because of its powerful and convenient fitting ability, it can solve the large and complex problems more effectively [5].

For the classification of red jujubes, there is no standard or even clear classification and grading criterion, and usually the classes are determined by people's experience. Deep learning is very good at learning the hidden pattern from labeled data set.

Therefore, this paper takes the classification of jujube as the research background and aims to combine deep learning techniques to construct a classification neural network structure that meets the real-time requirements in the current jujube system and improve the accuracy of jujube classification.

To build better deep learning architectures, this paper considers a multi-vision attention networks inspired by the multi-vision visual mechanism. The proposed model (presented in Fig. 4) incorporates some of the currently effective modules and connection methods in the convolutional neural network structure to complete the design of the model structure. From the first layer, the model not only simulates the effects of multiple eyes, but also abstracts the concept of the eye to the deep layer, so that it has parallel eyes from shallow to deep. The eye structure of our model mainly consists of a convolutional layer of $1\times1$ and $3\times3$, with a batch normalization (BN) layer and PReLU activation layer set in front of them [6]. The $1\times1$ convolution, which can be regarded as a bottleneck layer [7, 8], can compress the number of feature maps and reduce the computational cost. We named the eye as Visual Receptor. In order to make full use of the information seen by these Visual Receptors, we adopt a cross-layer connections similar to DenseNet [9] fuse each layer of the Visual Receptors, and features of the Visual Receptors of the same layer similar to the parallel structures of the Inception modules [10-13]. In addition, in order to further improve the representation ability, we include the channel type attention mechanism such as module of Squeeze-and-Excitation Networks [14] into our model. Benefits of our network architecture, we can obtain a powerful and real-time classification method to meet the requirement of red jujube grading system.

We evaluate the effectiveness of the attention mechanism and the multiple eye design respectively. And we also compare the performance of our model with others, such as DenseNet121 [9], Inceptionv3, v4 [10-13], and Squeeze-and-Excitation Networks [14]. The contributions of this paper are as follows:

---

[†] This work is supported by National Key R&D Program of China (No. 2018YFD0700300).

- A neural network model for real-time classification of red jujubes is constructed, which has not only high precision but also light structure and meets real-time requirements. Our model performances better in comparison with some of the more common classic models.
- A data set with four classes of red jujubes was built, such as invalid, rotten, wizened and normal.

This paper first introduces the current related works of convolutional neural network, and fruit classification methods. In section III we introduce the jujube grading system. The proposed architecture is presented in section IV. Section V contains the data set building for jujube classification and numerical results. We demonstrate the algorithm efficiency and give some discuss in Section VI. Section VII concludes the paper.

## II. Related Work

With widespread use of deep learning, many fruit and vegetable classification algorithms use a combination of traditional image algorithms and deep learning. For example: Sidehabi et al.[15] use the K-Means Clustering algorithm for cutting and then extracts RGB features into a simple neural network for the classification of Passiflora. Their experimental results are very good, but their data sets are too small, the generalization performance of the model is not well, and the artificial neural network is only a classification function. Zeng [16] uses image saliency to draw the object regions and convolutional neural network (CNN) VGG model to classify 26 types of fruits and vegetables. They built their own larger fruit and vegetable datasets, and their data was mostly from the web and daily shooting. Image saliency can be used to better adapt to complex background environments, but it depends on whether image saliency model can effectively focus on the foreground of the image. Khaing, Naung and Htut [17] propose an 8-layer simple CNN network to classify 30 fruit images on the FIDS30 fruit image data set. Based on the classic LeNet structure, they designed their own convolutional classification model, which has great optimization potential for the current rapid development of CNN.

Moreover, for jujube, some even use infrared spectroscopy information[18, 19]. A continuum of traditional image algorithms were used to construct a system for jujube maturity measurement and classification [19]. Their work is comprehensive, counting the jujube color histograms to determine categories, and using a thermal camera to detect defects. This undoubtedly increases the hardware cost for a model that uses only the camera. And G. Muhammad.[20] uses traditional image algorithms to extract local texture features and then classify them with support vector machines. Their workflow was classic before CNN prospered, and the quality of the model classification depends on the feature extraction. And their data set is too small. Then, some use simple neural network. A simple four-layer fully connected neural network is used to classify multiple types of jujubes by some physical attributes and appearance statistics in [21, 22]. The artificial neural network only plays a role as a classifier, and the effect of feature extraction greatly affects the classification result. So far, it is rare to use CNN to classify red jujubes, but CNN has achieved great success in image processing. Its local connectivity guarantee that a spatially local input pattern produces the strongest response by the learnt convolutional kernel. And its shared weights produce the property of translation invariance, and the amount of parameter variable can be reduced. These make CNN strong image feature extraction capabilities. So we consider building a model by CNN.

For the convolutional neural network (CNN), there are many classical models, such as AlexNet [5] (2012), VGG[23] (2014), Inception [10-13] (2014-2016), ResNet [8] (2015), DenseNet [9] (2017), and so on. However, for the jujube classification task, although these networks have strong feature extraction capabilities, they cannot meet the real-time requirements of the system. Some network models, such as ResNet18 [8], LeNet-5 [24], and SE-Net50[14], may meet the speed requirements, but their accuracy may not be satisfactory. So, we consider how to build a model with high accuracy and speed for jujube classification.

**Dense connectivity.** Based on the skip connection represented by ResNet [8], DenseNet [9] introduces dense connectivity which helps to better propagate features and losses. Dense connectivity can alleviate the vanishing-gradient problem, encourage different feature reuse, and substantially reduce the number of model parameters[9]. More importantly, DenseNet can directly be trained from scratch without pre-trained data. These characteristics of DenseNet are very practical for jujube classification since our classification problem has distinguish.

**Attention mechanisms.** Based on attention mechanisms, SENet's SE blocks can obtain significant performance improvements at slight computational cost [14] vis strengthening some neurons by weighting the activations channel-wisely. Another kind of visual attention is about weighting the activations spatially [25, 26]. Residual Attention Network can solve the problems that stacking spatial attention modules directly would cause the obvious performance drop[25], and its structure is more complicated than the channel-wise attention mechanism. In addition, the effectiveness of this method depends on the contrast of the foreground and background. It is more focused on location than on the entire object or a specific feature of an object.

**The parallel architecture.** Inspired by primate visual cortex, GoogLeNet proposed the Inception module, which is the improved utilization of the computing resources inside the network[10]. In recent years, Inception module has also integrated in other networks such as Inception-ResNet [13], which combines the merits of Inception and ResNet. Some methods [27-29] introduce the combination of DenseNet and Inception. Other networks that use parallel unit structures also perform well [27, 30].

**Activation.** ReLU[31] is a commonly used Rectifer with better gradient propagation and more efficient calculations, mainly for non-linearity transformations. But it is not differentiable at zero, and more seriously it sometimes leads to dying ReLU problem. Leaky ReLUs have a small negative slope can mitigate dying ReLU problem. Further, the slope of PReLU[6] in the negative region can be controlled by a coefficient of each neuron.

## III. Background

### 1. Jujube sorting system

As shown in Fig. 1, the image acquisition system used in this paper is a high-performance micro-diameter fruit and vegetable sorting machine developed by Jiangxi Reemoon Sorting Equipment Co., Ltd. and Institute of Microelectronics of Chinese Academy of Sciences. The machine vision part is mainly composed of high-resolution industrial cameras, photoelectric switch which is used to control image capture, LED light sources, and conveyor belts with rollers.

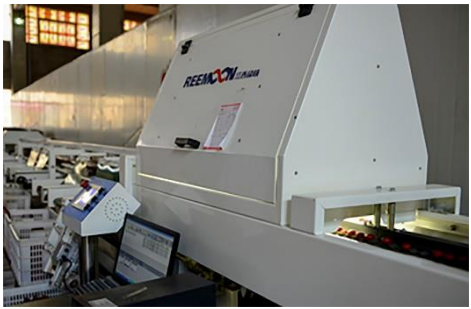

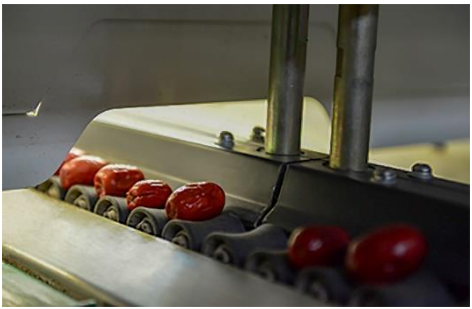

Fig. 1. The high-performance micro-diameter fruit and vegetable sorting machine. Among them, the above picture is the appearance of the entire system. The picture below shows the status of the jujube on the conveyor belt. The roller on the conveyor belt can rotate the red jujubes.

The camera resolution is $1280 \times 1024$ and its frame rate is 60 fps. In order to get the information of the whole surface of one jujube, the jujube rotates with the roller while the camera captures images. Thus, we get five images for one jujube. After applied preprocessing methods on every image, the jujube region is extracted (shown in Fig. 2). Then, we get the category for one jujube based on the five classification results for the five images. Furthermore, to meet the real-time requirement, the computational time for the jujube classification method should be less than 4ms per image.

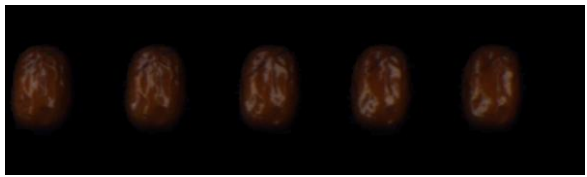

Fig. 2. From left to right: the five images captured while the jujube rotating.

From Fig. 2, we can observe that the illumination is low, since high illumination may introduce serious reflection which is not helpful for preprocessing and classification.

### 2. Data Set

As shown in Fig. 3, the red jujubes images are divided into four categories: invalid, rotten, wizened and normal. The invalid images are production of the preprocessing method used in the jujube sorting system. The other types of jujube are divided based on the appearance of the jujube surface and the feature of the sorting system.

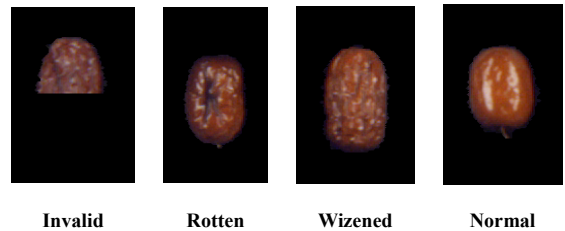

**Invalid        Rotten        Wizened        Normal**

Fig. 3. Four categories of red jujubes.

As presented in Table 1, our jujubes dataset is comprised of 23,735 $100 \times 200$ color images of 4 classes, with 21,380 training images and 2,355 test images. Before training, we hold out 1280 training images as the validation set.

Table 1 Basic situation of jujube data set

| Set | Invalid | Rotten | Wizened | Normal | Total |
| --- | --- | --- | --- | --- | --- |
| Training Set | 512 | 4942 | 7646 | 8280 | 21380 |
| Test Set | 55 | 550 | 850 | 900 | 2355 |
| Total | 567 | 5492 | 8496 | 9180 | 23735 |

## IV. The Proposed Model

In nature, most of the creatures with eyes have at least two eyes. A parietal eye, also called the third eye, is present in lizards, frogs, salamanders and other creatures. Most arachnids have eight eyes, and flies have compound eyes which are even composed of 4,000 small eyes (ommatidia). Mammals like humans mostly have a pair of eyes. Inspired by this fact, this paper proposes the multi-vision attention network (referred to as: MvANet) based on such a multi-eye visual mechanism. In our model, we call one eye a visual receptor. Furthermore the high-level visual receptor also treats the whole feature map of the low level layer as an eye which contains multiple cells.

### 1. Model architecture

We present the structure of MvANet in Fig. 4, which has multiple eyes in each layer of the backbone. The model is mainly composed of the following three parts: shallow feature extraction, deep feature extraction, and the final classification part. The structure of Multiple parallel visual receptors is called a visual layer.

Simple visual layer consists of multiple parallel convolution layers followed by a channel-wise module which assigns a certain weight to each feature channel and reduces the spatial size of the feature map by one-half. Then, the multiple outputs of the shallow feature extraction are concatenated together to merge the information.

The following multiple visual layers in visual receptors block are composed of multiple parallel visual receptors,

respectively. It should be noted that the size of parallel module in each visual layer can be arbitrary.

The output of the visual receptor of the same visual layer is concatenated together and then input to the next visual layer. Besides, dense connectivity is used between visual layers $1-l$. Finally, we use a single visual receptor to fuse all the features of the previous $l$ layers and compress the channels. Then, the channel-wise attention is followed to weight the channel of the feature map and further reduce the spatial size of the feature map.

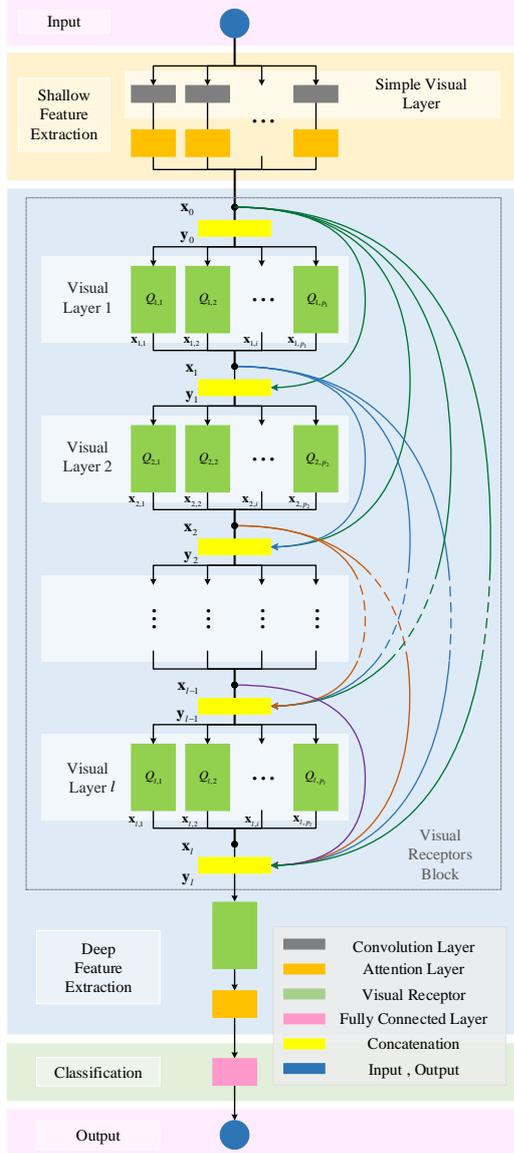

Fig. 4. MvANet Structure

The last part is the fully connected layer, which is mainly for outputting the final category. It uses the average pooling to process the input feature map into a vector and then make a fully connected output.

Next, each module and connection method will be described in detail.

## 2. Attention layer

The proposed model uses the channel-wise attention mechanism [14] which is equivalent to assign each channel a weight. We design the attention layer based on the transition layer of CliqueNet[32].

As presented in Fig. 5, a batch normalization layer [11] and a nonlinear transform layer PReLU [6] are added in front of the $1\times 1$ and $3\times 3$ convolution layer. We also use the PReLU in the first fully connected layer.

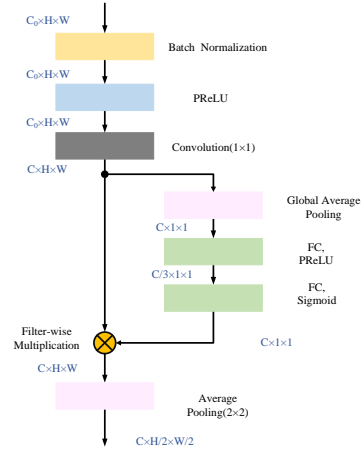

Fig. 5. $C$, $H$, and $W$ represent the channel, height, and width of the feature map, respectively. $C = a \cdot C_0$, where $a$ is a scaling factor, usually $0 < a \leq 1$.

## 3. Visual receptor

Fig. 6 shows visual receptor block which uses the bottleneck block to reduce the computational cost [33]. After the $1\times 1$ convolution reduced the parameter number, a $3\times 3$ convolution kernel is applied on the resulting feature maps. The $1\times 1$ convolution was proposed in Network in Network [7], and then wad combined in bottleneck building block in ResNet[8]. We add BN and PReLU before the two convolutions to introduce the regularization and enhance the representation ability of the network.

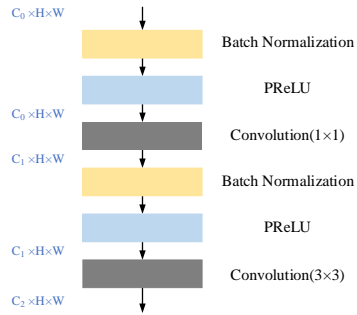

Fig. 6. A schema of visual receptor.

We have $C_1 = C_0 /(t+\alpha)$, where $t$ is the channel compression factor in the visual receptor. $\alpha = \frac{1}{2} p(p+1)/10$ which makes $t$ have an offset. $p$ represents the ocular number (The number of visual receptors in the visual layer.). $C_2 = k/p$, where k represents the number of output channels of the current visual layer.

## 4. Connection in visual receptors block

We use DenseNet connection[9] and Inception connection[10] to make full of use the information. As shown in Fig. 4, the input of each visual receptor in each layer is the concatenation of all the previous layers for visual receptor block. The dense connectivity is adopted between visual layers, and the visual receptors in visual layer uses a parallel approach similar to the Inception.

It is assumed that the input of $l$ th layer is denoted by $\mathbf{y}_{l-1}$ and the output is $\mathbf{x}_l$. A $\mathbf{x}_l$ consists of all visual receptors in $l$ th layer, and we define it as:

$$\mathbf{x}_l = \bigcup_{i=1}^{p_l} \mathbf{x}_{l,i} \triangleq [\mathbf{x}_{l,1}, \mathbf{x}_{l,2}, ..., \mathbf{x}_{l,p_l}] \quad (1)$$

where $p_l$ is the number of visual receptors in one layer. And the output of the $i$ th visual receptor of $l$ th layer is expressed as:

$$\mathbf{x}_{l,i} = Q_{l,i}(\bigcup_{j=0}^{l-1} \mathbf{x}_j) \quad (2)$$

where $Q$ is the operation of a visual receptor.

Therefore, the feature map for the output of $l$ th layer can be further expressed as:

$$\mathbf{x}_l = \bigcup_{i=1}^{p_l} Q_{l,i}(\bigcup_{j=0}^{l-1} \mathbf{x}_j) \quad (3)$$

Finally, we get the input $\mathbf{y}_l$ to the next module:

$$\mathbf{y}_l = \bigcup_{j=0}^{l} \mathbf{x}_j \quad (4)$$

## 5. Implementation details

We design the monocular, binocular, trinocular network architecture respectively, and also present a simplified version of the monocular structure, named as: MvANet-1, MvANet-2, MvANet-3 and MvANet-1-tiny. Their visual receptors block is set to three visual layers. The main difference between them is that the number of visual receptors in each visual layer is different ($p = 1, 2, 3, 1$, respectively). In addition, the number of convolution kernels in visual receptor is different. In order to reduce the computational burden, MvANet-1-tiny has light parameters mainly by setting lower values about $k_1$, $k_2$ and $k_3$. Table 2 gives the parameters of the convolutional layer used in the network architecture.

Table 2 Model parameters

| Model | MvANet-X | | MvANet-1-tiny | | MvANet-1 | | MvANet-2 | | MvANet-3 | |
|---|---|---|---|---|---|---|---|---|---|---|
| Parameters | $k_0, k_1, k_2, k_3, a_1, a_2, t, p$ | | $k_0=84, k_1=18, k_2=24, k_3=30,$ $a_1=1.0, a_2=0.8, t=1.7, p=1$ | | $k_0=84, k_1=96, k_2=138, k_3=192, t=1.5, a_1=0.7, a_2=0.6$ | | | | | |
| | | | | | $p=1$ | | $p=2$ | | $p=3$ | |
| Module | Output | Structural units | Output | Structural units | Output | Structural units | Output | Structural units | Output | Structural units |
| Simple Visual Layer | $[W \times H, C_0 = k_0/(a_1 \times p)] \times p$ | $[3 \times 3, k_0/(a_1 \times p) conv] \times p$ | $[32 \times 32, 84] \times 1$ | $[3 \times 3, 84 conv] \times 1$ | $[32 \times 32, 120] \times 1$ | $[3 \times 3, 120 conv] \times 1$ | $[32 \times 32, 60] \times 2$ | $[3 \times 3, 60 conv] \times 2$ | $[32 \times 32, 40] \times 3$ | $[3 \times 3, 40 conv] \times 3$ |
| Attention Layer A | $W/2 \times H/2, C_1 = k_0$ | $\begin{bmatrix} 1 \times 1, k_0/p conv \\ global\ avg\ pool \\ [k_0/p, k_0/(3p), k_0/p] fc \\ 2 \times 2\ avg\ pool \end{bmatrix}$ | $16 \times 16, 84$ | $\begin{bmatrix} 1 \times 1, 84 conv \\ global\ avg\ pool \\ [84, 28, 84] fc \\ 2 \times 2\ avg\ pool \end{bmatrix}$ | $16 \times 16, 84$ | $\begin{bmatrix} 1 \times 1, 84 conv \\ global\ avg\ pool \\ [84, 28, 84] fc \\ 2 \times 2\ avg\ pool \end{bmatrix}$ | $16 \times 16, 84$ | $\begin{bmatrix} 1 \times 1, 42 conv \\ global\ avg\ pool \\ [42, 14, 42] fc \\ 2 \times 2\ avg\ pool \end{bmatrix} \times 2$ | $16 \times 16, 84$ | $\begin{bmatrix} 1 \times 1, 28 conv \\ global\ avg\ pool \\ [28, 9, 28] fc \\ 2 \times 2\ avg\ pool \end{bmatrix} \times 3$ |
| Visual Layer 1 | $W/2 \times H/2, C_2 = C_1 + k_1$ | $\begin{bmatrix} 1 \times 1, C_1/(t+\alpha) conv\_1 \\ 3 \times 3, k_1/p conv\_2 \end{bmatrix} \times p$ | $16 \times 16, 102$ | $\begin{bmatrix} 1 \times 1, 46 conv\_1 \\ 3 \times 3, 18 conv\_2 \end{bmatrix} \times 1$ | $16 \times 16, 180$ | $\begin{bmatrix} 1 \times 1, 52 conv\_1 \\ 3 \times 3, 96 conv\_2 \end{bmatrix} \times 1$ | $16 \times 16, 180$ | $\begin{bmatrix} 1 \times 1, 46 conv\_1 \\ 3 \times 3, 48 conv\_2 \end{bmatrix} \times 2$ | $16 \times 16, 180$ | $\begin{bmatrix} 1 \times 1, 40 conv\_1 \\ 3 \times 3, 32 conv\_2 \end{bmatrix} \times 3$ |
| Visual Layer 2 | $W/2 \times H/2, C_3 = C_2 + k_2$ | $\begin{bmatrix} 1 \times 1, C_2/(t+\alpha) conv\_1 \\ 3 \times 3, k_2/p conv\_2 \end{bmatrix} \times p$ | $16 \times 16, 126$ | $\begin{bmatrix} 1 \times 1, 56 conv\_1 \\ 3 \times 3, 24 conv\_2 \end{bmatrix} \times 1$ | $16 \times 16, 318$ | $\begin{bmatrix} 1 \times 1, 112 conv\_1 \\ 3 \times 3, 138 conv\_2 \end{bmatrix} \times 1$ | $16 \times 16, 318$ | $\begin{bmatrix} 1 \times 1, 100 conv\_1 \\ 3 \times 3, 69 conv\_2 \end{bmatrix} \times 2$ | $16 \times 16, 318$ | $\begin{bmatrix} 1 \times 1, 85 conv\_1 \\ 3 \times 3, 46 conv\_2 \end{bmatrix} \times 3$ |
| Visual Layer 3 | $W/2 \times H/2, C_4 = C_3 + k_3$ | $\begin{bmatrix} 1 \times 1, C_3/(t+\alpha) conv\_1 \\ 3 \times 3, k_3/p conv\_2 \end{bmatrix} \times p$ | $16 \times 16, 156$ | $\begin{bmatrix} 1 \times 1, 70 conv\_1 \\ 3 \times 3, 30 conv\_2 \end{bmatrix} \times 1$ | $16 \times 16, 510$ | $\begin{bmatrix} 1 \times 1, 198 conv\_1 \\ 3 \times 3, 192 conv\_2 \end{bmatrix} \times 1$ | $16 \times 16, 510$ | $\begin{bmatrix} 1 \times 1, 176 conv\_1 \\ 3 \times 3, 96 conv\_2 \end{bmatrix} \times 2$ | $16 \times 16, 510$ | $\begin{bmatrix} 1 \times 1, 151 conv\_1 \\ 3 \times 3, 64 conv\_2 \end{bmatrix} \times 3$ |
| Visual Receptor | $W/2 \times H/2, C_5 = C_4/2$ | $\begin{bmatrix} 1 \times 1, C_4/(t+\alpha) conv\_1 \\ 3 \times 3, k_4/p conv\_2 \end{bmatrix} \times 1$ | $16 \times 16, 78$ | $\begin{bmatrix} 1 \times 1, 86 conv\_1 \\ 3 \times 3, 78 conv\_2 \end{bmatrix} \times 1$ | $16 \times 16, 255$ | $\begin{bmatrix} 1 \times 1, 318 conv\_1 \\ 3 \times 3, 255 conv\_2 \end{bmatrix} \times 1$ | $16 \times 16, 255$ | $\begin{bmatrix} 1 \times 1, 283 conv\_1 \\ 3 \times 3, 255 conv\_2 \end{bmatrix} \times 1$ | $16 \times 16, 255$ | $\begin{bmatrix} 1 \times 1, 242 conv\_1 \\ 3 \times 3, 255 conv\_2 \end{bmatrix} \times 1$ |
| Attention Layer B | $W/4 \times H/4, C_6 = a_2 \times C_5$ | $\begin{bmatrix} 1 \times 1, C_6 conv \\ global\ avg\ pool \\ [C_6, C_6/3, C_6] fc \\ 2 \times 2\ avg\ pool \end{bmatrix} \times 1$ | $8 \times 8, 62$ | $\begin{bmatrix} 1 \times 1, 62 conv \\ global\ avg\ pool \\ [62, 20, 62] fc \\ 2 \times 2\ avg\ pool \end{bmatrix} \times 1$ | $8 \times 8, 153$ | $\begin{bmatrix} 1 \times 1, 153 conv \\ global\ avg\ pool \\ [153, 51, 153] fc \\ 2 \times 2\ avg\ pool \end{bmatrix} \times 1$ | $8 \times 8, 153$ | $\begin{bmatrix} 1 \times 1, 153 conv \\ global\ avg\ pool \\ [153, 51, 153] fc \\ 2 \times 2\ avg\ pool \end{bmatrix} \times 1$ | $8 \times 8, 153$ | $\begin{bmatrix} 1 \times 1, 153 conv \\ global\ avg\ pool \\ [153, 51, 153] fc \\ 2 \times 2\ avg\ pool \end{bmatrix} \times 1$ |
| Classification Layer | $1 \times 1, C_7 = NumClasses$ | $\begin{bmatrix} global\ avg\ pool \\ [C_6, C_7] fc \end{bmatrix} \times 1$ | $1 \times 1, 4$ | $\begin{bmatrix} global\ avg\ pool \\ [62, 4] fc \end{bmatrix} \times 1$ | $1 \times 1, 4$ | $\begin{bmatrix} global\ avg\ pool \\ [153, 4] fc \end{bmatrix} \times 1$ | $1 \times 1, 4$ | $\begin{bmatrix} global\ avg\ pool \\ [153, 4] fc \end{bmatrix} \times 1$ | $1 \times 1, 4$ | $\begin{bmatrix} global\ avg\ pool \\ [153, 4] fc \end{bmatrix} \times 1$ |

\* $[\cdot] \times p$ represents p structures in parallel. And $\alpha = \frac{1}{2} p(p+1)/10$ .

MvANet-X is a generalized representation of the parameters. The $k_0$ represents the number of channels of the feature map input to the first visual layer. And $k_1$, $k_2$, $k_3$ in turn represent the output of the three visual layers. Then, $a_1$, $a_2$ represent the scaling factor in the attention layer. Next, $t$ is the compression factor in the visual receptor. Finally, $p$ represents the ocular number.

# V. Experiment

We perform the experiments on jujubes dataset and compare with state-of-the-art architectures. Then, the recall and precision are adopted to further comparison. To test the classification performance of the model in the system, we also test the accuracy of the individual red jujubes.

## 1. Training

All models were trained with mini-batch size 64 using Titan X cuDNN v6.0.21 with Intel Xeon E5-2683 v3 @ 2.00GHz. And we adopt some standard data augmentation scheme, such as mirroring and rotating. The images were normalized into [0,1] using mean values and standard deviations. Our models were mainly trained with images of resolution $32 \times 32$ from scratch. We also adopted the weight initialization method introduced by[34] .

The compared models were pre-trained on the ImageNet dataset, except for LeNet trained from scratch. Among them, because of the deeper models not effective supporting for $32 \times 32$ input, DenseNet and Inception models were trained with images of resolution $96 \times 96$, and the remaining models were trained with images of resolution $32 \times 32$. So, we also trained MvANet-3 by images of resolution $96 \times 96$ to conduct fair comparison. The optimization performance of amsgrad on dense connectivity model in our experiment was not as efficient as that of SGD [35], so our model and DenseNet121 used this optimizer with momentum 0.9. We set the initial learning rate at 0.1, and it is multiplied by 0.1 at 40%, 70% and 90% of the total number of training epochs.

And the optimization for other models were performed using the amsgrad [36] optimizer. In addition to we trained LeNet for 500 epochs, and the others for 300 epochs.

## 2. Result

We compare the experimental results of our models with some of state-of-the-art network architectures on our jujube dataset. The batch size is chosen to be 5 in testing, since we need to classify the five images of one red jujube each time.

We also normalized the images into [0,1] using mean values and standard deviations. And each of our architectures was trained 30 times from scratch. The mean and standard deviation of the estimated mean model skill are used as the criteria for accuracy evaluation, respectively written Avg Error and Std Error. At the same time, we also compare the lowest error rate from all the results, and it is written as Best Error. In Table 3, the best results in every part are marked in **boldface**.

Table 3　Comparison of MvANet with other network architecture.

| Model | Input resolution | Params (KB) | Avg Error(%) | Best Error(%) | Avg Time(ms/5 images) |
|---|---|---|---|---|---|
| DenseNet121 | 96×96 | 27,624 | - | 8.45 | 46.47 |
| Inception V3 | 96×96 | 85,354 | - | 8.75 | 34.21 |
| Inception V4 | 96×96 | 161,131 | - | 8.75 | 63.29 |
| Inception-ResNet v2 | 96×96 | 212,615 | - | 8.32 | 54.22 |
| MvANet-3 | 96×96 | **5,347** | 8.92 | **7.86** | **23.33** |
| LeNet | 32×32 | 3,758 | 17.31 | 15.29 | **3.04** |
| ResNet18 | 32×32 | 43,722 | - | 12.82 | 6.40 |
| SE-ResNet50 | 32×32 | 102,015 | - | 11.93 | 16.87 |
| NASNet-A Mobile[37] | 32×32 | 16,972 | - | 12.82 | 51.77 |
| MvANet-3 | 32×32 | 5,347 | **9.31** | **8.11** | 11.05 |
| MvANet-1-tiny | 32×32 | **479** | 10.69 | 9.85 | 5.26 |

The results in Table 3 indicate that MvANets are more efficiently than other architectures for utilizing parameters. Because of inheriting the characteristics of dense connectivity, MvANets also can substantially reduce the amount of parameters. Compared DenseNet with 121 convolutional layers, MvANet with 11 convolutional layers also achieves perfect performance for accuracy and speed. It is worth noting that Inception-ResNet v2, which has highest accuracy, has 40 times the network parameter size of MvANet-3. But MvANet-3 can reach its level of accuracy. And MvANet-1-tiny with 479KB parameters is less than a tenth of the parameters in MvANet-3.

**Accuracy.** Remarkably in Table 3, MvANet-3 achieves the best error of 7.86%. Although its average error of 8.92% is highest in 96×96 input resolution, it is just 0.17% higher than Inception V3 [12] and V4[13]. And it has the absolute superiority for the best error of 8.11% and the average error of 9.31% in 32×32 input resolution. The MvANet-3 models with different resolution shows that the training by higher resolution input indeed improve the accuracy. So, the compared model with higher resolution input has more advantages. However, it is more encourage that MvANet-3 with 32×32 input resolution achieves the best error of 8.11%, which meets the average level (8.57%) of average error with 96×96 input resolution. Our tiny model also has a good average error of 10.69%, which is 1.38% higher than MvANet-3 and 1.24% lower than SE-ResNet50.

**Computational cost.** For all 96×96 resolution input models, only MvANet-3 is available in real time of system. And MvANet-3 with 32×32 resolution input only consumes half of its time. Although the time consuming for LeNet and ResNet18 is very impressive, their accuracy is not satisfactory, even worse than MvANet-1-tiny with the same time consuming level.

We also compute precision and recall for each class and show the individual performance of our models and DenseNet, as shown in Table 4. All compared models have higher precision and recall on invalid class, but lower ones on rotten class. And Multi-ocular architectures have a better performance.

Table 4 MvANet-3 classification results for each type of red jujubes

| Model | Category | Recall | Precision |
|---|---|---|---|
| Dense Net | Invalid | **100.00** | 96.49 |
| | Rotten | 88.18 | 85.69 |
| | Wizened | 90.59 | 93.11 |
| | Normal | **93.00** | 92.49 |
| lMvANet-1 | Invalid | **100.00** | **100.00** |
| | Rotten | 90.00 | 85.35 |
| | Wizened | 90.82 | 93.46 |
| | Normal | 92.22 | 92.84 |
| MvANet-2 | Invalid | **100.00** | **100.00** |
| | Rotten | 90.18 | **86.41** |
| | Wizened | **91.53** | 93.74 |
| | Normal | 92.00 | 92.41 |
| MvANet-3 | Invalid | **100.00** | 98.21 |
| | Rotten | **91.64** | 85.14 |
| | Wizened | 89.77 | **94.08** |
| | Normal | 92.67 | **93.08** |

Finally, a new small test set consisting of 500 images corresponding to 100 red jujubes was constructed to test the accuracy of classifying individual red jujubes. We tested the model trained by 32×32 resolution input. In the table, MvANet still performs best, and MvANet-3 has the highest precision of 85%.

Table 5 Classification accuracy of 100 individual red jujubes

| Model | Precision (%) | Model | Precision (%) |
|---|---|---|---|
| LeNet | 76 | MvANet-1-tiny | 82 |
| ResNet18 | 78 | MvANet-1 | 82 |
| SE_ResNet50 | 79 | MvANet-2 | 84 |
| NASNet-A Mobile | 78 | MvANet-3 | 85 |

## VI. Discussion

We empirically demonstrate attention mechanism and multi-vision mechanism in MvANet and show the robustness to noisy labels for our model. Based on current experiments, the future work will also be discussed.

### 1. Attention mechanism

To verify the effectiveness of the attention mechanism, we refer to MvANet-1-tiny, which details in Table 2. The global average pooling and two fully connected layer (as shown in Fig. 5) are removed to become the transition layer, and such a network architecture is named MvTNet-1. So as shown in Fig. 7, the other modules of MvTNet-1 are consistent with MvANet-1-tiny except for the two transition layers.

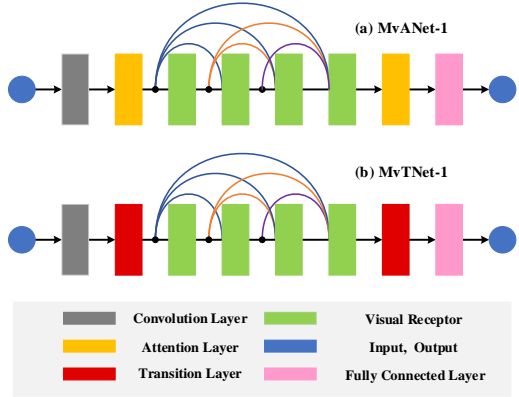

Fig. 7. MvANet-1-tiny and MvTNet-1 architecture comparison. MvTNet-1 only lacks the attention mechanism module (a global average pooling layer and two fully connected layer) than MvANet-1-tiny.

As shown in Table 6, the average error of MvANet-1-tiny is lower than MvTNet-1① 0.569%. And the best error of the MvANet-1-tiny is also 0.892% lower than MvTNet-1①. But it takes a little more time.

Table 6 The effectiveness of the attention layer.

| Model | Params (KB) | Avg Error(%) | Std Error | Best Error(%) | Avg Time(ms/5 images) |
|---|---|---|---|---|---|
| MvANet-1-tiny | 479 | 10.685 | 0.00361 | 9.851 | 5.2563 |
| MvTNet-1① | 445 | 11.254 | 0.00329 | 10.743 | 4.4133 |
| MvTNet-1② | 487 | 11.231 | 0.00378 | 10.318 | 4.4037 |

However, the Table 6 also shows that MvANet-1 has a higher parameters (34KB) than MvTNet-1① who removes the attention mechanism module. In order to eliminate the possibility of high parameters to improve classification accuracy, we increase the size of parameters in MvTNet-1① to get MvTNet-1② (Set $k_3 = 32$). Now, although the parameter size of MvTNet-1② is 8KB higher than MvANet-1-tiny, the average error and the best error are still not as good as MvANet-1-tiny. So, this further proves that our architecture added the attention mechanism module is effective.

### 2. Multi-vision mechanism

We trained MvANets with the different ocular numbers (Set $p=1$, $p=2$ and $p=3$, respectively), which details in Table 2. Then, the results, the most noticeable trend from the first part of Table 7, shows that the average error drops from 9.532% to 9.529% and finally to 9.305% and the number of parameters decreases from 6,224KB, down 5,870KB to 5,347KB as the ocular numbers increasing. The best error that reduce from 8.62% to 8.408% and to 8.11% also has the similar trend. This suggests the multi-ocular of MvANet is effective. Although increasing the number of oculars consumes more time, the most time-consuming MvANet-3, which takes 11.045ms/5 images, still meets the real-time requirements.

Table 7 The effectiveness of multiple visual mechanism.

| Model | Params (KB) | Avg Error(%) | Std Error | Best Error(%) | Avg Time(ms/5 images) |
|---|---|---|---|---|---|
| MvANet-1 | 6,224 | 9.532 | 0.00390 | 8.620 | 6.831 |
| MvANet-2 | 5,870 | 9.529 | 0.00488 | 8.408 | 9.017 |
| MvANet-3 | 5,347 | **9.305** | 0.00497 | **8.110** | 11.045 |

In addition, we selected a representative image from each category as input and visualized the output of each visual layer and the shallow feature extraction for MvANet-3, respectively. As shown in Fig. 8, the feature map output by each visual receptor (VR I, VR II and VR III) is different for the same input. This suggests that the function of the visual receptors in the same visual layer is different.

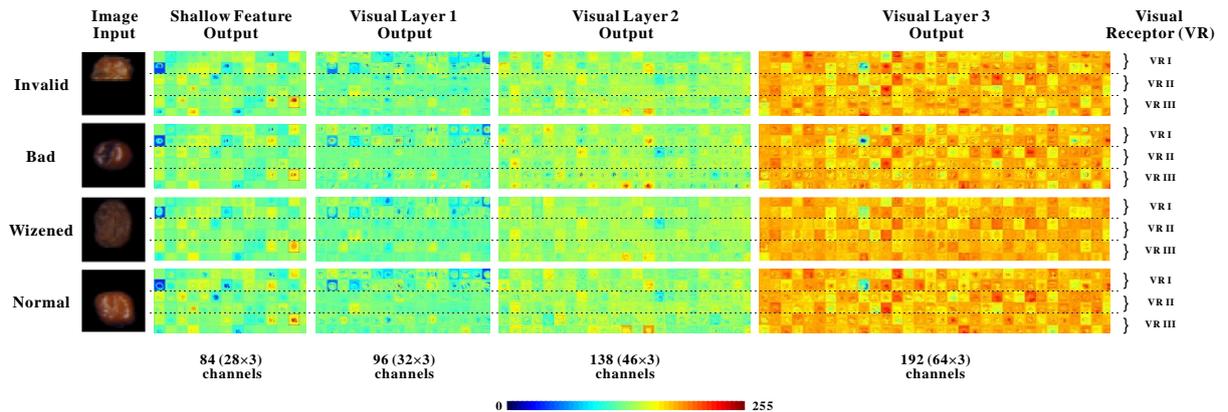

Fig. 8. Visualization of the outputted feature map from each visual layer and shallow feature extraction (MvANet-3). The output of each visual receptor in the same layer is different with the same input. And as deeper with the number of layers, the value of the visual receptor output gradually increases.

### 3. Robustness to noisy labels

Our model can distinguish some noisy tags, as show in Fig. 9. Both #0012640 and #0014696 have distinct black flaws. There is obvious mildew spot in the upper right corner of #0022570. And the blemish of #0002191 with cracked skin also is conspicuous. Finally, the last one is wizened. Even though the model was trained using noisy labels whereas mistaken labels are not as harmful to the performance. So, our models also have robustness to noisy labels.

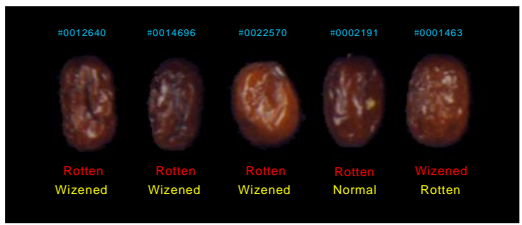

Fig. 9. The red label is the result of the network output, and the yellow is the manual label. Although these tags are wrong, the network made the right judgment.

## 4. Future work

As shown in Fig. 10, there is a trend that the jujube is more and more wizened from right to left. The surface of red jujubes has no defects in the first row, but the last row has. And a small vertical flaw appears on jujubes in the second row. So, it is difficult to set a category criterion, especially for two adjacent jujubes in the same row, a jujube with two classification characteristics and a jujube with small defects.

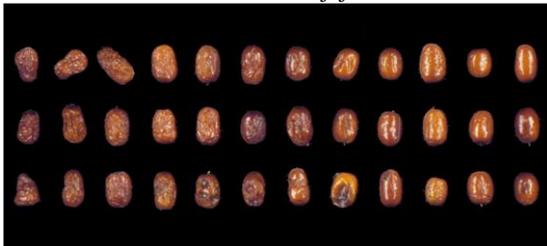

Fig. 10. Continuous wizened changes are difficult for manual labeling.

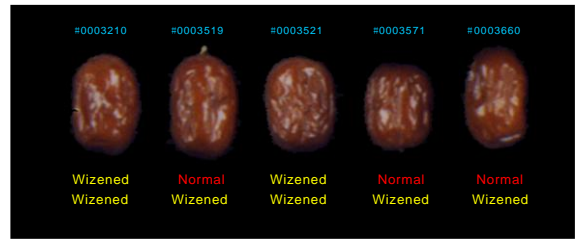

Fig. 11. The result of the $96 \times 96$ resolution MvANet-3 output corresponds to the red jujube image and label. The row with red word is the output from the model and the last row is the correct label.

As shown in Fig. 11, it is not too terrible to judge some of the ambiguous red jujubes by ours model. And different customers may have different classification requirements. We will further set reasonable classification criteria and think about multi-label[38].

In addition, the models designed in the project will also be considered for testing on public datasets. To further verify whether the structure is versatile for other similar data sets.

## VII. Conclusion

We proposed the Multi-vision Attention Network (MvANet), which is a new convolutional network architecture. And we also built a data set with four classes of red jujubes. In the experiments, MvANet yield better results with the addition of attention module and multiple visual mechanism.

MvANet-3 achieved the accuracy of 91.89%, which reached the average level (91.43%) of deeper state-of-the-art network architectures that can not meet the real-time requirements. But it can, it only consumes one-third of the time, which is the system can reserve for the classification algorithm. We also designed MvANet-1-tiny, a simpler architecture, and the time it consumed to classify was only half of MvANet-3. Compared to some state-of-the-art small network architectures, it also has absolute superiority for the accuracy of 90.15%.